\documentclass[10pt,twocolumn,letterpaper]{article}

\usepackage[pagenumbers]{wacv} 

\usepackage{graphicx}
\usepackage{amsmath}
\usepackage{amssymb}
\usepackage{booktabs}
\usepackage{xcolor}
\usepackage{adjustbox}
\usepackage{xfrac} 
\usepackage[pagebackref,breaklinks,colorlinks]{hyperref}
\usepackage[accsupp]{axessibility}

\usepackage[capitalize]{cleveref}
\crefname{section}{Sec.}{Secs.}
\Crefname{section}{Section}{Sections}
\Crefname{table}{Table}{Tables}
\crefname{table}{Tab.}{Tabs.}

\def\wacvPaperID{} 
\def\confName{WACV}
\def\confYear{2024}
\newcommand{\paperTitle}[0]{Synthesizing Anyone, Anywhere, in Any Pose}

\newcommand{\appExpDetails}[0]{Appendix \textcolor{red}{A}\xspace}
\newcommand{\IM}[0]{\bar{I}}
\newcommand{\EX}[0]{\mathbb{E}}
\newcommand{\Pl}[0]{P_\ell}

\newcommand{\fidC}[0]{FID$_{\text{CLIP}}$\xspace}
\newcommand{\methodName}[0]{TriA-GAN\xspace}
\newcommand{\appFDH}[0]{Appendix  \textcolor{red}{B}\xspace}
\newcommand{\appFnetQualitative}[0]{Appendix  \textcolor{red}{C}  \xspace}
\newcommand{\appRandomQualitative}[0]{Appendix  \textcolor{red}{D}  \xspace}
\newcommand{\appSGGANAnonymizationComparison}[0]{Appendix  \textcolor{red}{E}  \xspace}

\definecolor{blue}{HTML}{377EB8}
\definecolor{orange}{HTML}{FF7F00}

\begin{document}

\title{\paperTitle}

\newcommand{\authorskip}{\hspace{10mm}}
\author{
   H{\aa}kon Hukkel{\aa}s  \authorskip Frank Lindseth \\
Norwegian University of Science and Technology \\
{\tt\small hakon.hukkelas@ntnu.no}
}
\twocolumn[{%
\renewcommand\twocolumn[1][]{#1}%
\maketitle
\begin{center}
    \centering
    \captionsetup{type=figure}
    \includegraphics[width=\textwidth]{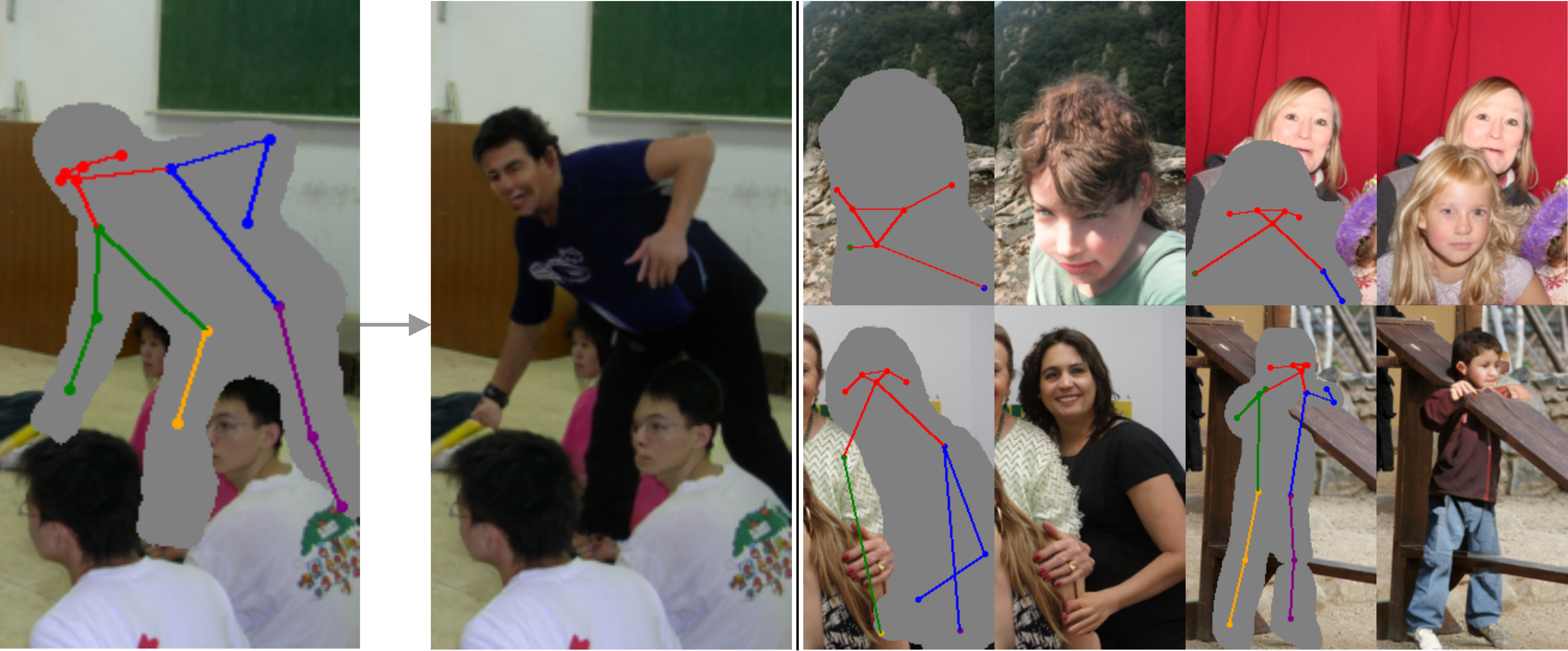}
    \captionof{figure}{
        \methodName can synthesize realistic human figures given a masked image and a sparse set of keypoints.
    }
    \label{fig:task_figure}
\end{center}%
}]

\begin{abstract}
  
  We address the task of in-the-wild human figure synthesis, where the primary goal is to synthesize a full body given any region in any image.
In-the-wild human figure synthesis has long been a challenging and under-explored task, where current methods struggle to handle extreme poses, occluding objects, and complex backgrounds.

Our main contribution is TriA-GAN, a keypoint-guided GAN that can synthesize Anyone, Anywhere, in Any given pose.
Key to our method is projected GANs combined with a well-crafted training strategy, where our simple generator architecture can successfully handle the challenges of in-the-wild full-body synthesis.
We show that TriA-GAN significantly improves over previous in-the-wild full-body synthesis methods, all while requiring less conditional information for synthesis (keypoints \vs DensePose).
Finally, we show that the latent space of TriA-GAN is compatible with standard unconditional editing techniques, enabling text-guided editing of generated human figures.

\end{abstract}
\section{Introduction}
Given any image with a missing region, can you imagine a human appearance fitting into it? 
If there is a football next to the missing region, does your imaginary person change?
This is a fascinating and difficult problem because countless possible solutions could fit the context.
We refer to this task as in-the-wild human figure synthesis.
Addressing this problem requires a complex understanding of human appearances and how they vary based on different environmental conditions, viewpoints, poses, and sizes of the missing region.
Such a system would have widespread applications in content creation, fashion \cite{Liu2016a}, or even for anonymization purposes \cite{Hukkelas2022a}.

Human figure synthesis is a well-established research field with many high-level goals.
However, \emph{in-the-wild} human figure synthesis is a difficult and under-explored task.
Previous methods focus on simpler tasks, such as transferring a known appearance into a given pose \cite{Balakrishnan2018,Chan2019}, transferring garments \cite{Han2018,Sarkar2020}, or full-body synthesis into a plain background \cite{Fruhstuck2022InsetGAN}.
Often they disregard the key difficulties of in-the-wild-synthesis, such as overlapping objects, partial bodies, complex backgrounds, and extreme poses.
In fact, recent studies filter out these difficult cases from their dataset to improve synthesis quality \cite{Fruhstuck2022InsetGAN,Fu2022StyleGANHuman}.
To the best of our knowledge, only a handful of research studies have tackled these challenges, with a focus on full-body synthesis for anonymization \cite{Hukkelas2022a,Hukkelas2022}
\footnote{Note that other studies address similar tasks \cite{Sun2018person,Song2021}, but they focus on simpler datasets (\ie Market1501 \cite{Zheng2015}, DeepFasion \cite{Liu2016a}) with few overlapping/occluding objects.}.
While previous methods \cite{Hukkelas2022a} generate visually pleasing results,  they heavily rely on DensePose estimation and struggle in complex scenarios.
In addition, the generated images are hard to edit \cite{Hukkelas2022a}.

A key issue of current methods for in-the-wild human figure synthesis  is their reliance on DensePose annotations \cite{Hukkelas2022a,Hukkelas2022}.
The available datasets with such annotations are either limited in size \cite{AlpGuler2018,Hukkelas2022} or automatically annotated \cite{Hukkelas2022a}.
We argue that this reliance constrain these methods, either by overfitting on small datasets \cite{Hukkelas2022} or by the numerous annotation errors arising from DensePose \cite{Hukkelas2022a}.

This paper explores full-body synthesis conditioned on sparse 2d-keypoints, eliminating the need for expensive DensePose annotations.
However, this  increases  the modeling complexity considerably, as the generative model must now infer both the body's texture \emph{and} its structure.
We find that current GANs \cite{Hukkelas2022a} struggle to synthesize realistic human figures without DensePose correspondences.

Our contributions address the challenge of scaling up GANs to handle in-the-wild full-body synthesis without DensePose correspondences.
Key to our method is replacing the conventional GAN discriminator with Projected GANs \cite{Sauer21PG}.
By combining Projected GANs with a thoughtfully designed training strategy, our method can generate coherent bodies with visually pleasing textures.

Our contributions can be summarized as follows.
First, we adapt Projected GANs \cite{Sauer21PG} for image inpainting (\cref{sec:pggan_IP}), and propose a novel mask-aware patch discriminator (\cref{sec:patch_discriminator}).
Secondly, we investigate the representational power of pre-trained feature networks used by the discriminator (\cref{sec:fnets}).
Our experiments reflect that the previously used classification networks \cite{Sauer21PG,Sauer22XL} are poorly suited for discriminating human figures.
Instead, we use a combination of self-supervised feature networks for the discriminator, which significantly improves sample quality.
Finally, we propose a progressive training technique for U-Net \cite{Ronneberger2015} architectures (\cref{sec:PG}), enabling us to easily scale up to high resolutions and larger model sizes.

Our contributions culminate into a new state-of-the-art for in-the-wild human figure synthesis.
As far as we know, our approach is the first to generate nearly photorealistic humans without DensePose annotations while effectively dealing with extreme poses, complex backgrounds, partial bodies, and occlusions.
Source code and appendix: \url{http://github.com/hukkelas/deep_privacy2/}.

\section{Related Work}

\subsection{Full-body Human Synthesis}
Synthesizing human bodies has a range of applications, and previous studies have a large variety of high-level goals.
We categorize human synthesis into \emph{transfer-based} and \emph{synthesis-based} models.
\emph{Transfer-based} methods transfers a source appearance (or garment \cite{Han2018,Sarkar2020}) into a new pose \cite{Balakrishnan2018,Li2019,Ma2017,Pumarola2018,Sarkar2020,Si2018}, motion \cite{Chan2019} or scene \cite{Siarohin2018}.
While some of these methods are applicable for in-the-wild human figure synthesis \cite{Siarohin2018,Yang2022}, they require a source appearance that limits the synthesized identities to a texture bank or an image dataset of appearances.
In contrast, our method can directly synthesize novel identities.
For the latter goal, \emph{synthesis-based} methods can synthesize the appearance either conditioned on a pose \cite{Sun2018person,Song2021,Yang223dHumanGAN}, scene \cite{Esser2018,Hukkelas2022a,Hukkelas2022}, or unconditionally \cite{Fruhstuck2022InsetGAN,Chaudhuri2021,Fu2022StyleGANHuman}.
Several of these methods are applicable for in-the-wild human synthesis \cite{Hukkelas2022a,Sun2018person,Song2021}, but they are limited to low-resolution \cite{Esser2018,Sun2018person}, struggle to handle complex backgrounds \cite{Fruhstuck2022InsetGAN,Song2021}, and only a few handles overlapping objects \cite{Hukkelas2022a,Hukkelas2022}.

Independent of the goal, most methods use a form of pose information to enhance synthesis quality through DensePose annotations \cite{Hukkelas2022a,Hukkelas2022,Neverova2018,Sarkar2020}, semantic segmentations \cite{Chaudhuri2021,Song2021,Yang2022}, sparse keypoints \cite{Balakrishnan2018,Chan2019,Esser2018,Han2018,Li2019,Ma2017,Sun2018person,Pumarola2018,Si2018,Siarohin2018,Yang2022}, or a 3d pose of the body \cite{Lassner2017,Yang223dHumanGAN}.

Previous studies primarily focus on GAN-based methods, but recent studies have employed diffusion models \cite{sohl2015deep} for human figure synthesis \cite{Jiang2022}.
Our work focuses on GANs as they offer fast sampling of high-quality images.

\subsection{Generative Adversarial Networks}
Generative Adversarial Networks \cite{Goodfellow2014} (GANs) have long been a leading generative model for a range of full-body synthesis tasks.
GANs are notoriously difficult to train, and a notable research focus has been on achieving stable training of the generator, where different techniques such as novel objectives \cite{Arjovsky2017}, architectures \cite{Karnewar2019a,Karras21Alias,Karras2018,Karras2019b}, training strategies \cite{Karras2017}, and regularization \cite{Gulrajani2017,Miyato2018} has been proposed to improve stability and synthesis quality.
Recently introduced Projected GANs \cite{Sauer21PG} use pre-trained feature networks for the discriminator to reduce training time and improve image quality, which was later extended for high-resolution image synthesis on the ImageNet \cite{Deng2010ImageNet} dataset \cite{Sauer22XL}.
We continue this line of research, where we adapt projected GANs for conditional synthesis.

\subsection{Image Inpainting}
Image inpainting \cite{bertalmio2000image} aims to complete missing regions in natural images.
Unlike general image inpainting, we complete missing regions that contain human figures appearing at random regions in natural images.
GANs have long been the leading methodology for free-form image inpainting \cite{Pathak2016,Yu2018GConv}, where most prior work focuses on architectural changes to the generator.
For example, to handle missing values \cite{Hukkelas21IP,Liu2018PConv,Yu2018GConv}, generate higher resolution \cite{Yi2020}, utilize auxiliary information \cite{Jo_2019,Lahiri_2020,Nazeri2019EdgeConnect}, or improve the receptive field via attention mechanisms \cite{Yu2018ContextualAttention} or fourier convolutions \cite{Suvorov22LAMA}.
Previous methods adapt a traditional GAN discriminator, often patch discriminators \cite{Isola2017,Liu2019,Ren_2019,Xiong2019,Yu2018GConv}, combined with perceptual image similarity losses \cite{Liu2019,Ren_2019} and pixel-wise $l_1$ loss \cite{Ren_2019,Xiong2019}.
As far as we know, we are the first to adapt Projected GANs \cite{Sauer21PG} for image inpainting, where we exclusively train on the adversarial objective.

\begin{figure*}
    \includegraphics[width=\textwidth]{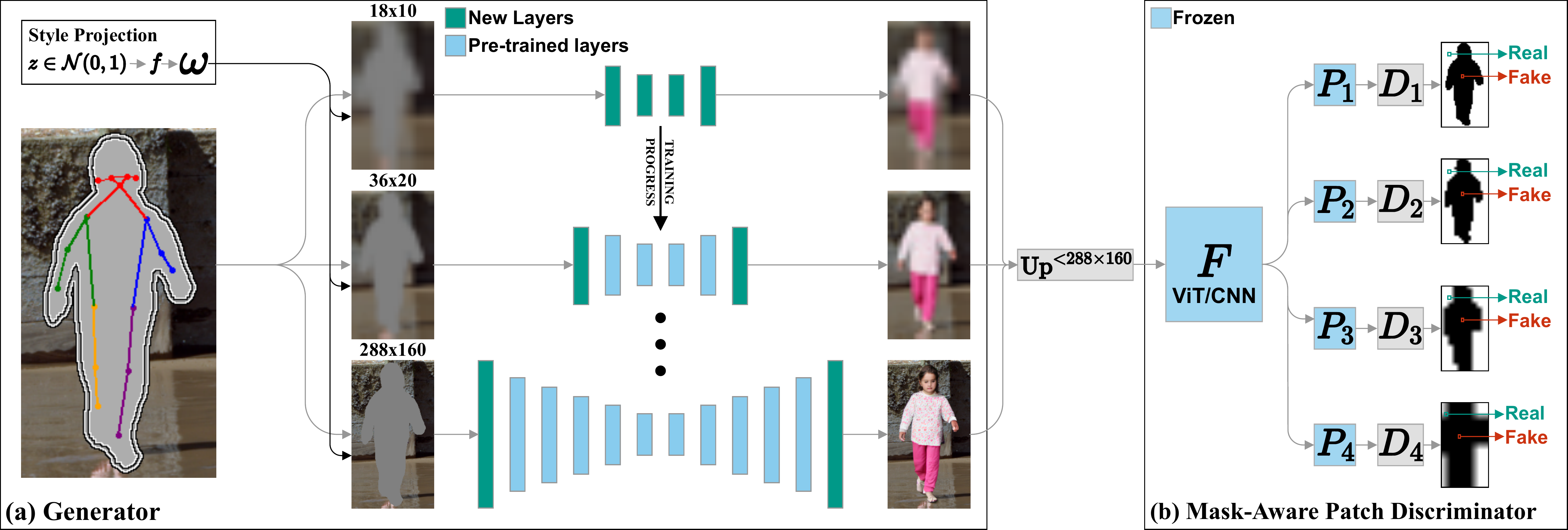}
    \caption{
        (a) Our generator fills in the missing region given 17 keypoints.
        The generator layers employ adaptive instance normalization to condition the generator on $\omega$, where $\omega$ is the output of the style mapping network.
        Config D\&E is trained progressively starting at $18 \times 10$ resolution, then increased by adding layers to the start/end of the encoder/decoder.
        Note that all layers remain trainable throughout training.
        (b) For each feature network $F$, we use four shallow patch discriminators operating its features (with different spatial resolutions), where each feature is projected through random differentiable operations ($P_1$-$P_4$).
        Given the projected features, each discriminator predicts if a given patch corresponds to a real or fake image region.
    }
    \label{fig:method_overview}
  \end{figure*}
  
\section{\methodName \xspace- A Keypoint-Guided GAN}
In this section, we gradually introduce changes to improve synthesis quality (\cref{tab:iterative}).
\textbf{Config A} (\cref{sec:pggan_IP}) starts with a StyleGAN-based \cite{Karras2018} U-Net \cite{Ronneberger2015} architecture, similar to the architecture used in \cite{Hukkelas2022a}, trained with Projected GANs \cite{Sauer22XL} using EfficientNet-Lite0 \cite{Tan2019}.
\textbf{Config B} introduces our Mask-Aware Discriminator objective (\cref{sec:patch_discriminator}), and
\textbf{Config C} replaces EfficientNet-lite0 with ViT-L16$_\text{MAE}$ and RN50$_\text{CLIP}$ (\cref{sec:fnets}).
\textbf{Config D} introduces our progressive training technique (\cref{sec:PG}) and finally, \textbf{Config E} increases the generator model size.
To reduce training time, we ablate our method on low-resolution images ($72 \times 40$).
Finally, \cref{sec:large_gans} increases the resolution to $288 \times 160$.
\appExpDetails includes experimental and architecture details.

\paragraph{Problem Formulation}
We formulate in-the-wild full-body synthesis as an image inpainting task.
Our goal is to complete the missing regions of a corrupted image $\IM = I \odot M$, where $I$ is the ground truth image, $M$ is the mask indicating missing regions ($M_i=1$ for known pixels and 0 for missing), and $\odot$ is element-wise multiplication.
To improve synthesis quality, we condition the generator on 17 keypoints following the COCO \cite{Lin2014COCO} keypoint format

\paragraph{Dataset} We conduct our experiments on the FDH dataset \cite{Hukkelas2022a}.
The FDH dataset is a large unfiltered dataset, where models trained on FDH adapt well to in-the-wild settings \cite{Hukkelas2022a}.
The dataset consists of 1.87M training images and 30K validation images.
Each image includes a single human figure as the subject, but the same image can include several individuals.
Each image is annotated with a 2d keypoint annotation, a segmentation mask indicating the human to be inpainted, and pixel-to-surface correspondences (\ie surface of a T-shaped 3D body).
Note that \methodName does not use pixel-to-surface correspondences.

We find that a large amount of the keypoint annotations in the FDH dataset are incorrect.
Thus, we automatically re-annotate all images with ViTPose \cite{Xu22VITPose} (see \appFDH).

\paragraph{Pose Representation}
We represent keypoints as a one-hot encoded spatial map, specifically $P \in \{0, 1\}^{K \times H \times W}$ where $K=17$ and $P_{k,y,x}=1$ for keypoint $k$  with location $(x, y)$ and $P$ is 0 otherwise.
In addition, we include a spatial map ($S$) drawing the human skeleton.
Specifically, the spatial map $S \in \{0, 1\}^{6 \times H \times W}$ is one-hot encoded into 6 categories, where lines connect closeby joints in the body, separated into 6 classes (left/right arm/leg, torso, head).
The one-hot encoded pose and the skeleton map are concatenated with the input image of the generator.

\begin{table}[t]
    \caption{
        Iterative development of our method. Each addition is added on top of the previous.
        Config A-C are trained until the discriminator has observed 50M images.
        }
    \label{tab:iterative}
    \centering
    \begin{adjustbox}{max width=\linewidth}
    \begin{tabular}{lcccc}
        \toprule
        Configuration & FID $\downarrow$ & \fidC $\downarrow$ & PPL $\downarrow$ & OKS $\uparrow$\\
        \midrule
        \textbf{A}: \hspace{.5em} Baseline & 1.73 & 1.74 & 55.8 & 0.916\\
        \textbf{B}: + Mask-Aware Discriminator & 1.65 & 1.63 & 52.8 & 0.912\\
        \textbf{C}: + Improved Feature Nets & 1.79 & 0.47 & 49.2 & 0.951 \\
        \textbf{D}: + Progressive Growing & 1.66 & 0.40 & 52.0 & \textbf{0.954}\\
        \textbf{E}:  + Larger G (62M $\rightarrow$ 110M) & \textbf{1.62} & \textbf{0.30} & \textbf{52.0} & 0.948\\
        \bottomrule
    \end{tabular}
    \end{adjustbox}

\end{table}

\paragraph{Evaluating Sample Quality}
We evaluate sample quality with Fr\'echet Inception Distance (FID) \cite{Heusel2017FID} and \fidC 
\footnote{ImageNet-FID scores images containing ImageNet objects higher and is insensive to faces \cite{kynkaanniemi2022role}. These issues are diminished with \fidC, where we use features from a CLIP \cite{Radford2021CLIP} pre-trained ViT-B/32.}. 
Additionally, we report latent disentanglement via Perceptual Path Length (PPL) \cite{Karras2018}, which correlates with consistency and stability of shapes \cite{Karras2019b}.

Furthermore, we introduce a new metric for assessing the sample quality of generated human figures, namely Object Keypoint similarity (\emph{OKS}), that compares the generated pose to the ground truth keypoints.
The motivation behind this metric is to obtain a metric that is not influenced by the feature network used by the discriminator.
Projected GANs \cite{Sauer21PG} are known to achieve artificially good scores on feature-based metrics \cite{kynkaanniemi2022role}, which makes it challenging to make quantitative comparisons across different types of feature networks.
This is evident from our experiments, where Config B (which uses ImageNet features for the discriminator) generates severely more corrupted images than Config E but still achieves a similar ImageNet FID.

Object Keypoint Similarity (\emph{OKS}) is calculated by predicting keypoints with ViTPose \cite{Xu22VITPose}, 
then computing the OKS to the ground truth keypoints following COCO \cite{Lin2014COCO}.
Compared to direct Euclidean distance, OKS considers that predicted keypoints can deviate slightly from the ground truth keypoints, where the acceptable deviation varies for different keypoints (\eg the shoulder keypoint can deviate more than the eye keypoint).

\subsection{Projected GANs for Image Inpainting }
\label{sec:pggan_IP}

Projected GANs \cite{Sauer21PG} employ pre-trained feature networks to discriminate between real and fake images.
Given an image $I$, the adversarial objective is formulated as
\begin{equation}
\begin{split}
\min_G \max_{D_\ell} \sum_{\ell \in \mathcal{L}} \EX_{I \sim p_{data}} \left[ \log \left( D_\ell \left( \Pl \left( I \right) \right) \right) \right] + \\
\EX_{z\sim p_z} \left[ \log \left( 1 - D_\ell \left( \Pl \left( G \left( z, \bar{I} \right) \right) \right) \right) \right],    
\end{split}
\label{eq:uncond_adversarial_game}
\end{equation}
where $\{D_\ell\}$ is a set of independent discriminators operating on its feature projector $\Pl$.
Each projector  is frozen during training and consists of a pre-trained feature network $F$, where features from $F$ are randomly projected with differentiable operations.
For the baseline (\textbf{Config A}), we use EfficientNet-Lite0 \cite{Tan2019} as $F$ following \cite{Sauer21PG}, which we later revisit in \Cref{sec:fnets}.
For each discriminator $D_\ell$, we adopt a patch discriminator architecture, described in \Cref{sec:patch_discriminator}.

\Cref{eq:uncond_adversarial_game} does not enforce consistency between the condition ($\bar{I}$) and the generated image, yielding a  generator that learns to completely ignore $\IM$ in practice.
Thus, we enforce condition consistency by masking the output of the generator.
Specifically, we set $G(z, \bar{I}) = \tilde{I} \odot (1-M) + \bar{I} \odot M$, where $\tilde{I}$ is the output of the last layer in $G$.

\subsubsection{Stabilizing the Generator}
Naively adopting projected GANs for image inpainting is unstable to train and prone to mode collapse early in training.
This originates from the generator struggling to keep up with the pre-trained discriminator, where the discriminator overpowers the generator early in training.
To improve stability, we introduce several modifications to the adversarial setup.
First, we blur images inputted to the discriminator  at the start of training, where the blur is linearly faded over 4M images.
The long blur prevents the discriminator from focusing on the high-frequency edges caused by the masking of the generator output.
Previous methods apply discriminator blurring over the first 200k images \cite{Karras21Alias,Sauer22XL}, whereas we find it beneficial to significantly increase this period.
Furthermore, the U-net architecture injects the latent code ($z$) via a mapping network and style modulation following StyleGAN2 \cite{Karras2019b}.
We set the mapping network to 2 layers and reduce the dimensionality of $z$ to 64, following \cite{Sauer22XL}.
Furthermore, we scale residual skip connections by $\sfrac{1}{\sqrt2}$ (similar to \cite{Karras2019b}), and $\sfrac{1}{\sqrt3}$ for skip connections where residual U-net connections are present.
Finally, we use instance normalization instead of weight demodulation \cite{Karras2019b}, as we find it more stable to train.

\subsection{Mask-Aware Patch Discriminator}
\label{sec:patch_discriminator}

\begin{figure*}
    \begin{subfigure}{0.25\textwidth}
        \includegraphics[width=\textwidth]{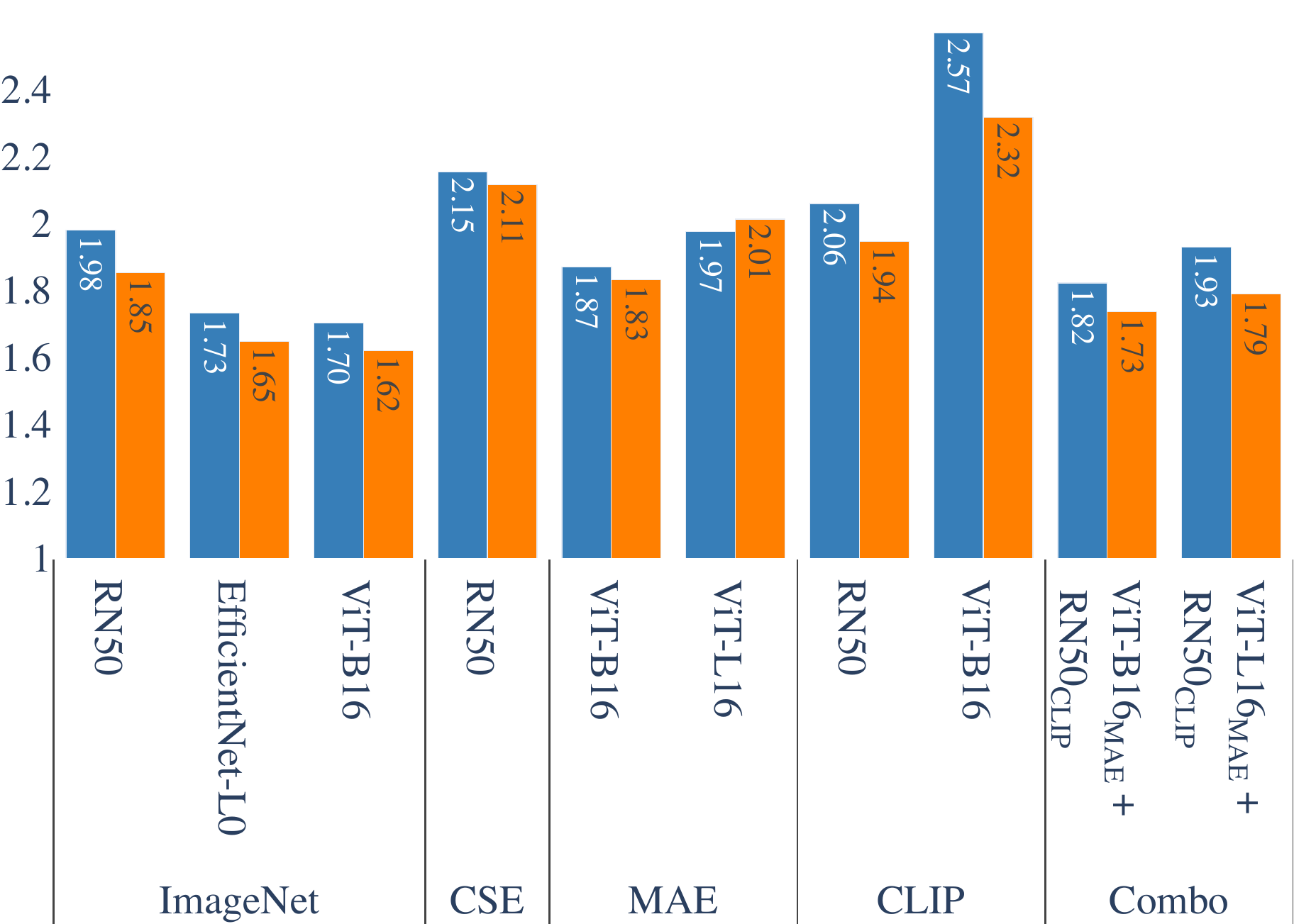}
        \caption{FID $\downarrow$}
    \end{subfigure}%
    \begin{subfigure}{0.25\textwidth}
        \includegraphics[width=\textwidth]{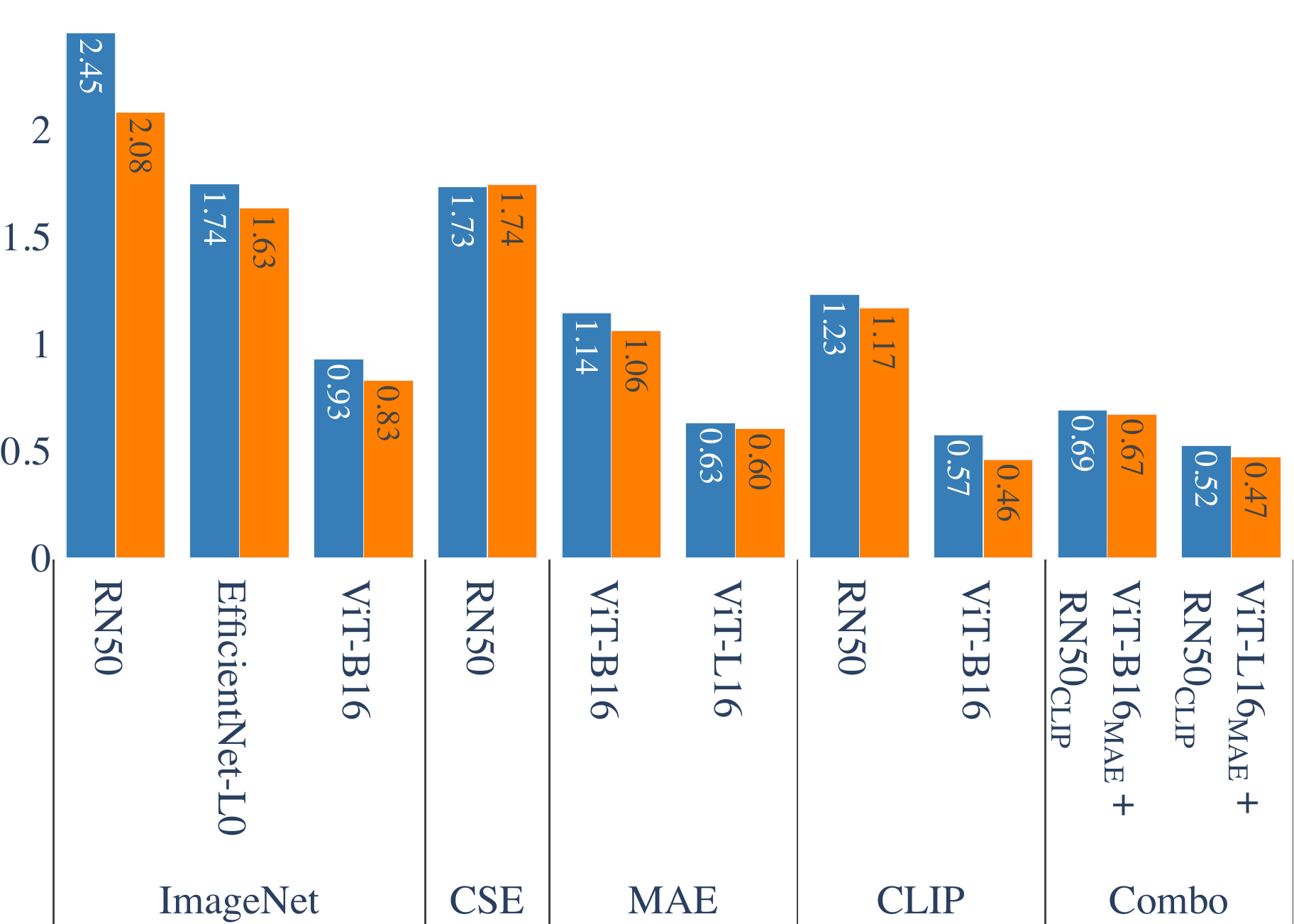}
        \caption{\fidC $\downarrow$}
    \end{subfigure}
    \begin{subfigure}{0.25\textwidth}
        \includegraphics[width=\textwidth]{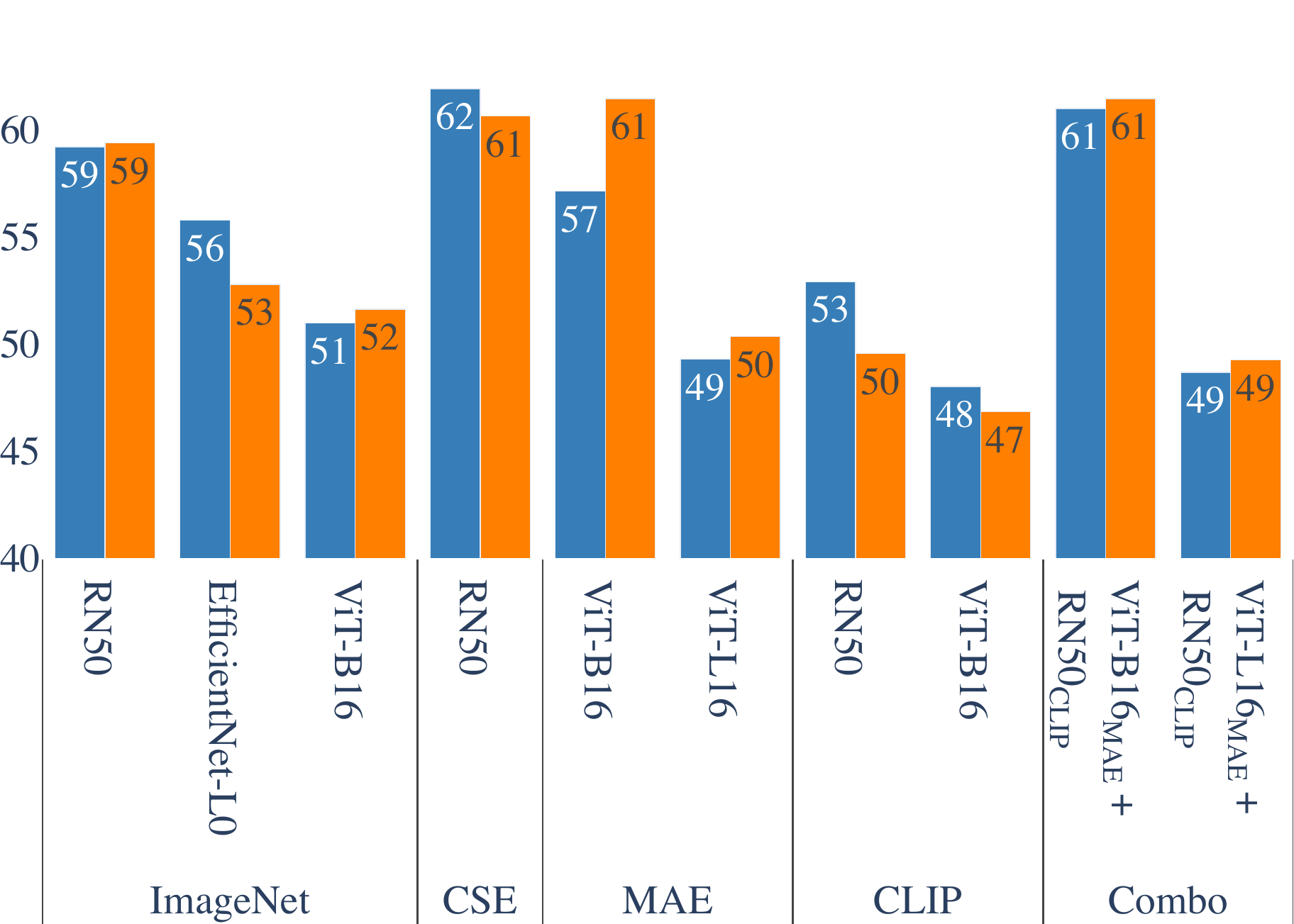}    
        \caption{PPL $\downarrow$}
    \end{subfigure}%
    \begin{subfigure}{0.25\textwidth}
        \includegraphics[width=\textwidth]{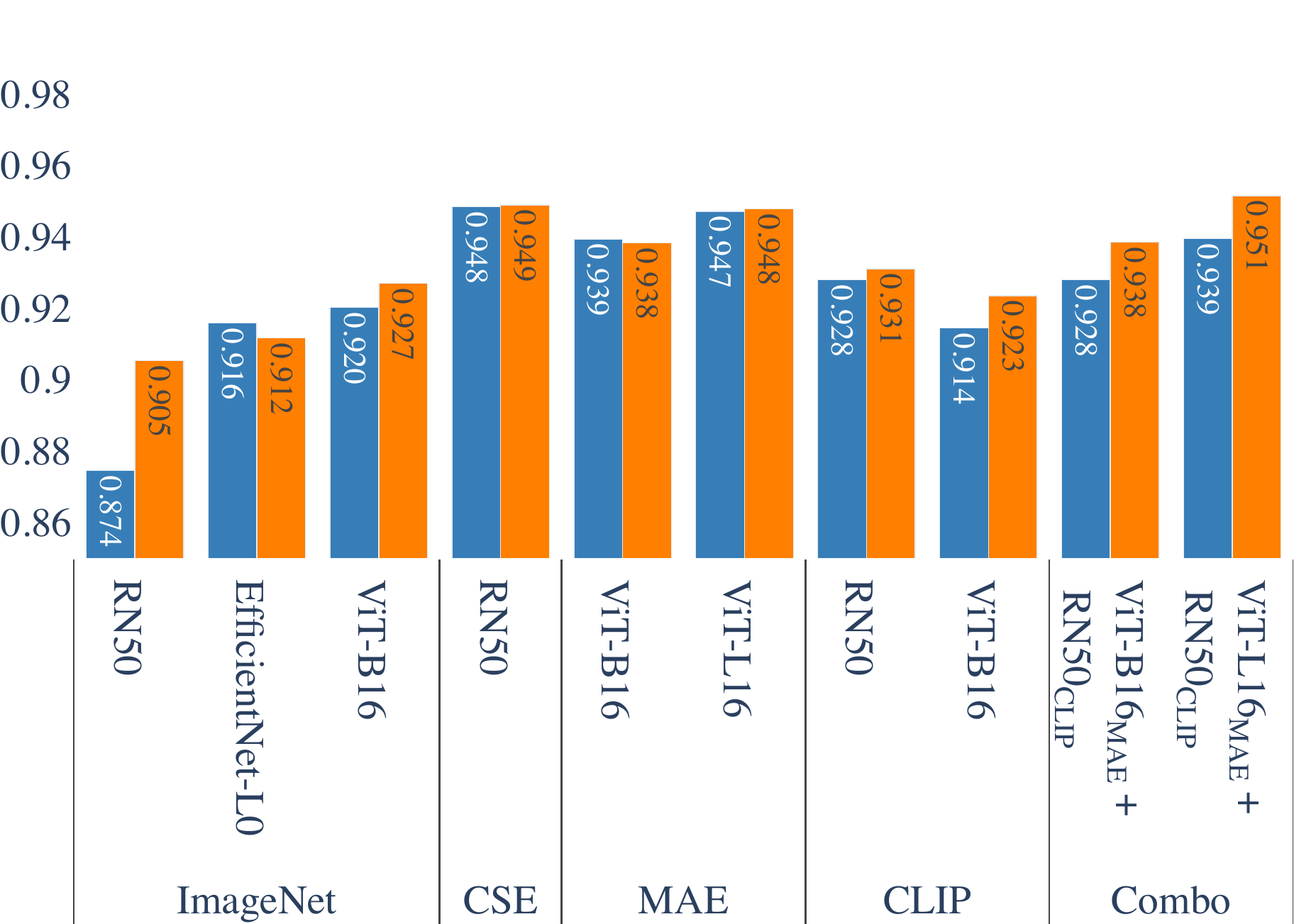}
        \caption{OKS $\uparrow$}
    \end{subfigure}%
    
    \caption{
        Comparison of different feature networks with the \textcolor{blue}{standard projected GAN objective} (\cref{eq:uncond_adversarial_game}) and \textcolor{orange}{mask-aware discriminator objective} (\cref{eq:patch_adversarial_game}).
        All models are trained until the discriminator has observed 50M images.}
    \label{fig:fnet_comparison}
\end{figure*}
Projected GANs \cite{Sauer21PG,Sauer22XL} adapt four shallow discriminators operating on different feature projections ($P_\ell$) with different spatial resolutions.
Each discriminator output logits at the same resolution ($4 \times 4$).
In contrast, we find patch discriminators to work better for the image inpainting task, where each discriminator tries to classify local patches instead of the global image.
Specifically, each $D_\ell$ (inputting features from the projection $P_\ell$) consists of three convolutions, where the output of $D_\ell$ is half the spatial resolution of $P_\ell$.
We find that replacing the discriminator from \cite{Sauer21PG} with a patch discriminator substantially improves performance.

Patch discriminators are widely adapted for image inpainting \cite{Suvorov22LAMA,Yu2018GConv,yu2019region,Zeng2020}.
Typically, each patch is classified as belonging to the class of the original image, such that all patches corresponding to a real image are classified as real.
However, this introduces ambiguity for the image inpainting task, as certain features (\eg shallow features from CNNs) might exclusively depend on real pixels even though the image is fake due to a limited receptive field.
Thus, we propose a mask-aware discriminator objective, where the discriminator's patches are categorized as belonging to the real or fake class based on whether they correspond to a real or fake region in the image.
The new objective is given by
\begin{equation}
\begin{split}
\min_G \max_{D_\ell}  \sum_{\ell \in \mathcal{L}}   \EX_{I \sim p_{data}} \left[ \log \left( D_\ell \left( \Pl \left( I \right) \right) \right) \right] +  \\
\EX_{z\sim p_z} \left[ \sum_{y}^{H_\ell} \sum_{x}^{W_\ell}   M_\ell^{y,x} \cdot \log \left( D_{\ell}^{y,x} \left(\Pl \left(G\left(z, \bar{I} \right) \right) \right) \right) + \right. \\
\left( 1-M_\ell^{y,x} \right) \cdot  \log(1 - D_\ell^{y,x}(\Pl(G(z, \bar{I})))) \bigg] ,
\end{split}
\label{eq:patch_adversarial_game}
\end{equation}
where $D_\ell \in \mathbf{R}^{H_\ell \times W_\ell}$, and $M_\ell$ is downsampled from $M$ to $H_\ell \times W_\ell$ via min-pooling.

\Cref{eq:patch_adversarial_game} removes the ambiguous classification of patches due to global class allocation, which provides more detailed and spatial coherent responses to the generator.
Furthermore, it introduces an auxiliary task to the discriminator, which is known to improve synthesis quality \cite{Odena2017}.
In our case, the auxiliary task is to spatially segment the region that corresponds to the generated area.

\Cref{fig:fnet_comparison} confirms that \Cref{eq:patch_adversarial_game} improves image quality (FID/\fidC) and OKS across a range of feature networks.
This includes feature networks with different pre-training tasks and architectures (CNNs and ViTs).
Similar segmentation discriminators have been explored before for other tasks \cite{Schonfeld20UNet,Sushko22OASIS,Yang223dHumanGAN}.
Our work further validate that this concept generalizes to extremely shallow discriminator architectures leveraging pre-trained feature networks, independent on the feature network used as $F$.

\subsection{Discriminative Feature Networks for Human Synthesis}
\begin{figure*}
    \centering
    \begin{subfigure}{.25\textwidth}
        \includegraphics[width=\textwidth]{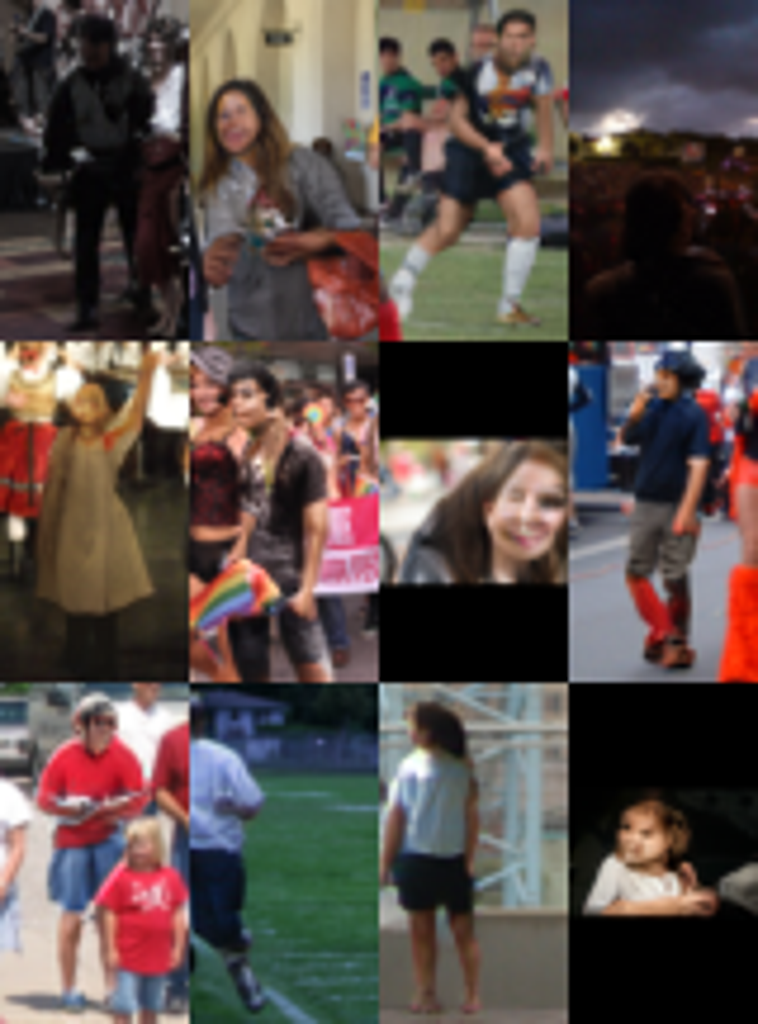}
        \caption{ImageNet (ViT-B/16)}
    \end{subfigure}%
    \begin{subfigure}{.25\textwidth}
        \includegraphics[width=\textwidth]{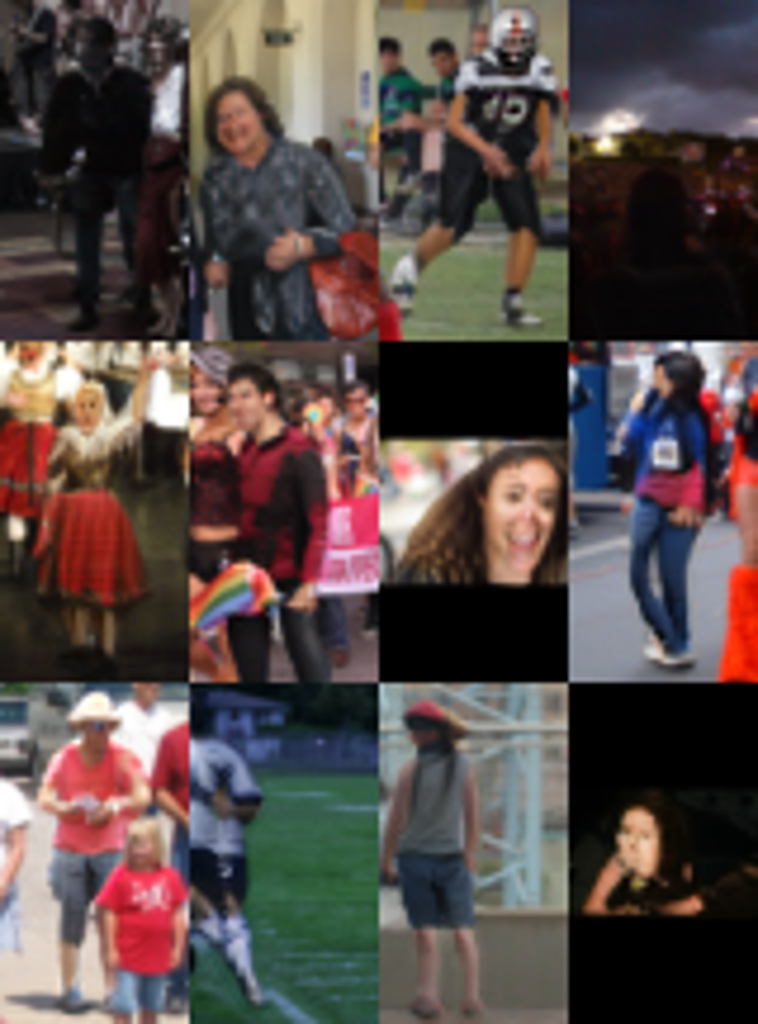}
        \caption{CLIP (ViT-B/16)}
    \end{subfigure}%
    \begin{subfigure}{.25\textwidth}
        \includegraphics[width=\textwidth]{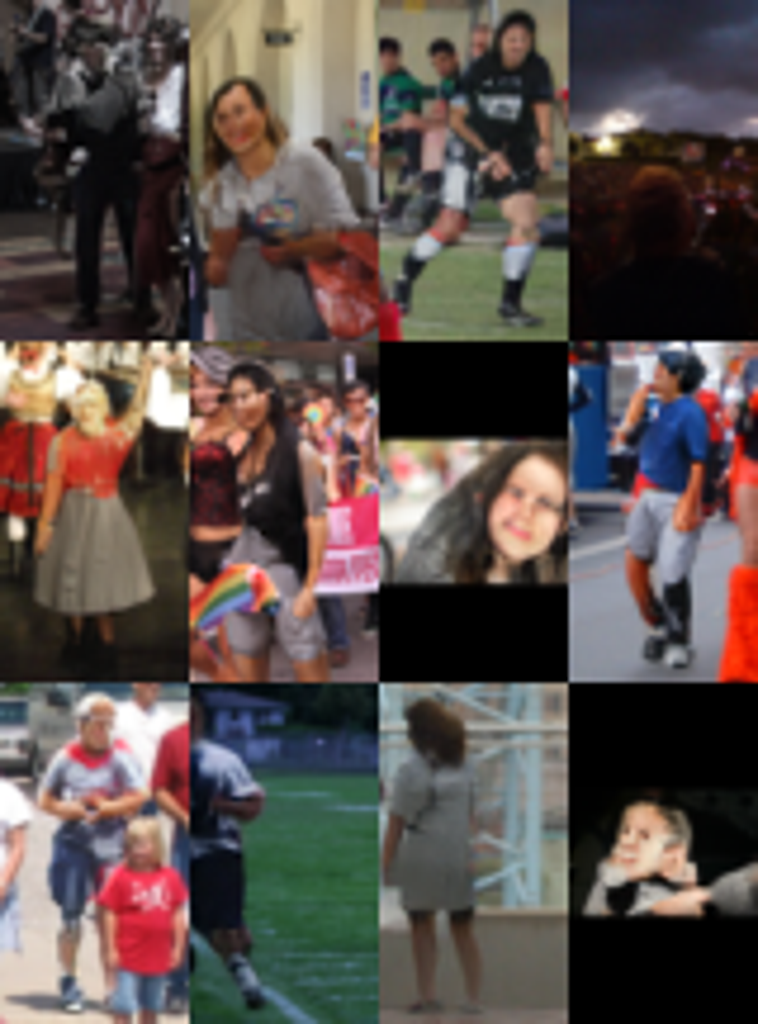}
        \caption{MAE (ViT-B/16)}
    \end{subfigure}%
    \begin{subfigure}{.25\textwidth}
        \includegraphics[width=\textwidth]{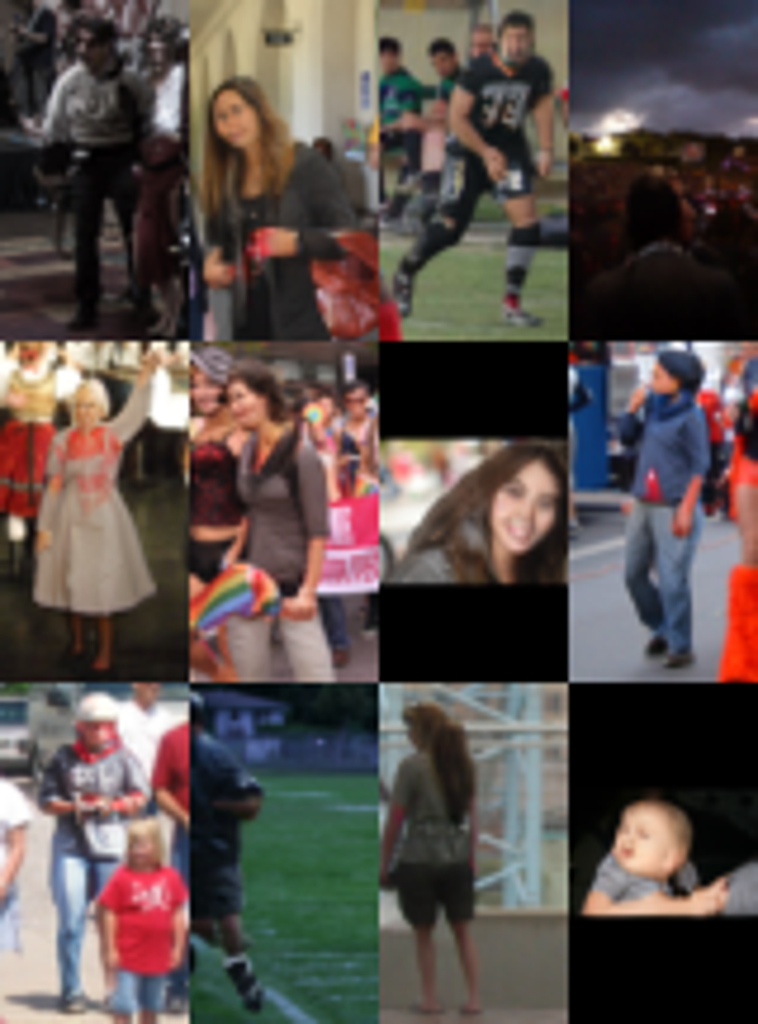}
        \caption{ViT-L/16$_\text{MAE}$ + RN50$_\text{CLIP}$}
    \end{subfigure}
    \caption{Qualitative comparison of various feature networks used for the discriminator. It is worth noting that these examples are not curated but selected from the first 12 images from the validation set.}
    \label{fig:fnet_qualitative_comparison}
\end{figure*}

\label{sec:fnets}
GANs have historically generated impressive results for aligned human synthesis, especially on the FFHQ \cite{Karras2018} and CelebA-HQ \cite{Karras2017,Liu2015CelebA} datasets.
However, projected GANs are known to generate artifacts for face synthesis on FFHQ \cite{Sauer21PG} and struggle to generate realistic images of unaligned humans \cite{Sauer22XL} \footnote{See the appendix in \cite{Sauer22XL}.}.
We find that the poor human synthesis quality originates from an invariance in the pre-trained feature space used by the discriminator.
Earlier work \cite{Sauer21PG,Sauer22XL} has utilized pre-trained ImageNet \cite{Deng2010ImageNet} classification networks.
These feature networks learn feature representations for the sole goal of classification; mapping an image to the top-1 class.
Hence, they learn to ignore features that are irrelevant to the goal of classification.
While this invariance benefits image classification, we find it  to hurt discriminative representation for human synthesis.

We explore different feature networks (including variants of CNNS/ViTs) with widely different pre-training tasks for the discriminator.
Specifically, \Cref{fig:fnet_comparison} ablate the following feature nets with the following pre-training tasks:
\begin{itemize}
    \item \textbf{IN}: ImageNet Classification: ResNet50 (\textbf{RN50}), \textbf{ViT-B16} (DeIT variant), EfficientNet-Lite0 (\textbf{EN-L0}).
    \item \textbf{CLIP}: Contrastive Language Image Pre-training \cite{Radford2021CLIP}:  \textbf{RN50}, \textbf{ViT-B16}.
    \item \textbf{MAE}: Masked Autoencoders \cite{He2022MAE}: \textbf{ViT-B16}, \textbf{ViT-L16}.
    \item \textbf{CSE}: DensePose estimation \cite{Neverova2020}: ResNet50 (\textbf{RN50}).
\end{itemize}
We refer to each model as $\textit{architecture}_\textit{task}$, \eg $\text{RN50}_\text{CLIP}$ refers to ResNet-50 with CLIP pre-trained weights.
Directly selecting the best feature network from standard generative metrics (FID/\fidC) is ambiguous, as projected GANs are known to achieve unnatural high scores on feature-based metrics \cite{kynkaanniemi2022role}.
We find that ImageNet models achieve unnatural high FID due to matching pre-training tasks, and  ViT scores better on \fidC due to matching architecture \footnote{\fidC is calculated from features of ViT-B/32 following \cite{kynkaanniemi2022role}.}.

Independent of the architecture, we observe that all ImageNet \cite{Deng2010ImageNet} models generate highly corrupted faces, illustrated in \Cref{fig:fnet_qualitative_comparison}.
This is most likely due to the invariance of facial descriptors in these feature networks, a phenomenon that has also been observed in \cite{kynkaanniemi2022role}.
Note that \appFnetQualitative includes comparison for all networks in \Cref{fig:fnet_comparison}.

From the results in \Cref{fig:fnet_comparison}, \textbf{Config C} replaces EfficientNet-Lite0 with ViT-L16$_\text{MAE}$ and RN50$_\text{CLIP}$.
The motivation for pairing these networks is to exploit features with completely different architectures and pre-training tasks.
In addition, these networks scores among the best \wrt OKS, \fidC, and PPL.
Finally, RN50$_\text{CLIP}$ supplements ViT well, as RN50 operates on the original aspect ratio ($288\times 160$), whereas ViT is fixed to $224 \times 224$
\footnote{ViT input resolution is set to $224 \times 224$ for all models, as ViT features are less robust to changes in resolution from the training resolution.}.

\subsection{Progressive Growing}
\label{sec:PG}
Progressive training \cite{Karras2017} is known to improve training stability of GANs and was recently re-introduced for unconditional synthesis with projected GANs \cite{Sauer22XL}.
StyleGAN-XL \cite{Sauer22XL} first trains at $16 \times 16$ resolution, then increases the resolution by adding new layers to the end of the decoder.
Note that StyleGAN-XL freezes already trained layers and the style network when training the next stage.

We adopt a straightforward extension to the image-to-image translation case, where we progressively train the U-net architecture by adding layers to the start/end of the encoder and decoder, respectively (see \cref{fig:method_overview}).
We observe that adding new blocks to the start of the encoder leads to training instability as it results in significant changes to the input of already-trained layers.
To mitigate this, we introduce LayerScale \cite{Touvron21} for each residual block with an initial value of $10^{-5}$ to lessen the contribution of new blocks.
Furthermore, we include output skip connections following \cite{Karras2018}.
Unlike StyleGAN-XL, we avoid freezing any blocks during training as the computational benefit is minimal, given that we need to calculate gradients for layers at the beginning of the encoder.
Introducing these changes substantially improves the final image quality (\textbf{Config D})

We note that we experimented with more advanced techniques for progressive training, such as cascaded U-nets \cite{Ho2022}, or assymetric training of the encoder/decoder (\ie start with a full-resolution encoder and a low-resolution decoder).
However, we found that the straightforward progressive training technique was superior in terms of training time and final image quality.

\subsection{Scaling Up the Generator}
\label{sec:large_gans}
\textbf{Config E} double the number of residual blocks for each resolution in the encoder/decoder, resulting in 110.4M parameters in the generator compared to the previous 62.2M.
This model trains stable up to $288 \times 160$ resolution, which is the maximum resolution of the FDH dataset.

\section{Comparison to Surface-Guided GANs}

\begin{figure}
    \includegraphics*[width=.98\columnwidth]{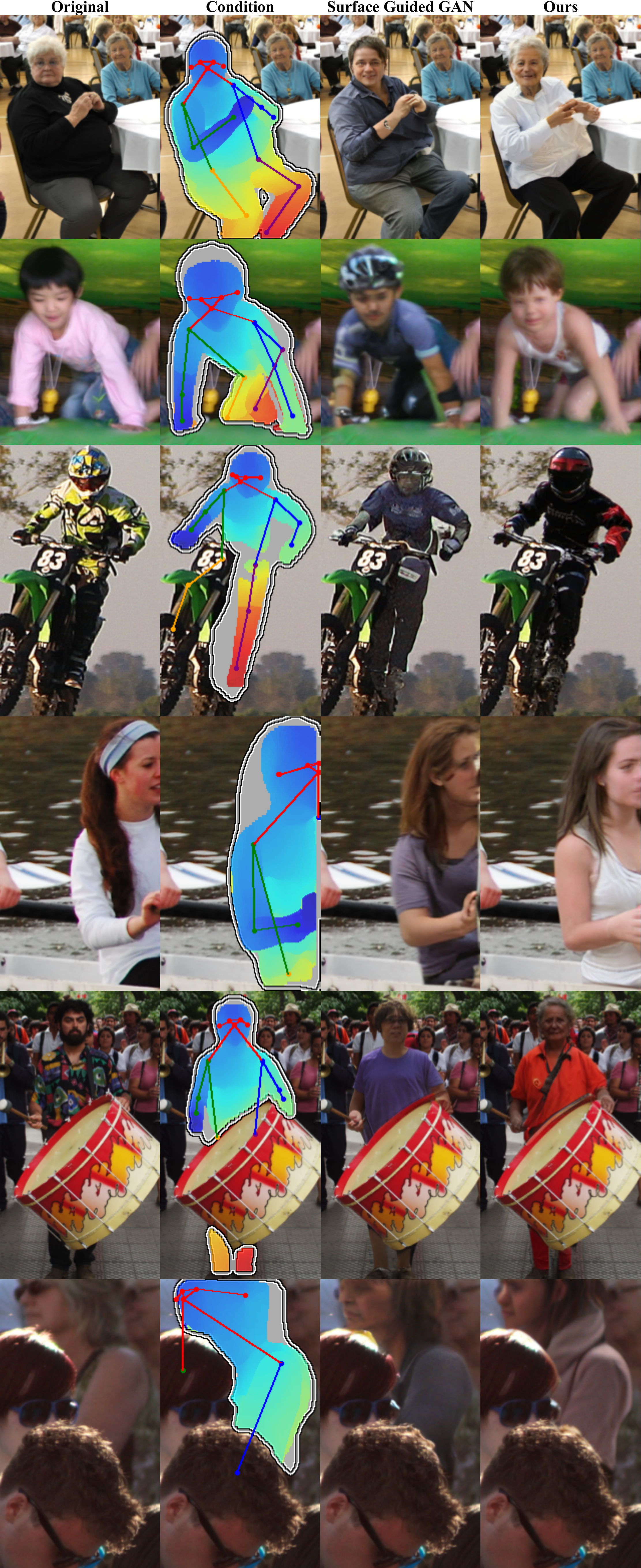}
    \caption{Curated examples comparing Surface Guided GAN \cite{Hukkelas2022a} to \methodName.
    Note that surface information is not used for \methodName (shown in blue-yellow tint).}
    \label{fig:sg_gan_vs_ours}
\end{figure}

\Cref{tab:sggan_vs_ours} compares \methodName to Surface Guided GANs (SG-GAN) \cite{Hukkelas2022} trained following DeepPrivacy2 \cite{Hukkelas2022a}, the current state-of-the-art for in-the-wild full-body synthesis.
\Cref{fig:task_figure} shows synthesis results with \methodName, and \Cref{fig:sg_gan_vs_ours} compares \methodName to SG-GAN.
\appRandomQualitative include randomly selected samples.

The main difference between \methodName and SG-GAN \cite{Hukkelas2022a} is the improved training strategy of \methodName, and the sparser conditional information (keypoints \vs dense surface correspondences).
\methodName improves at handling overlapping objects, partial bodies (\eg intersection with image edges), and synthesis of texture (\eg hair, clothing).
Furthermore, \methodName improves at context handling, \eg inferring that an elderly lady is likely to sit at the table (top row, \cref{fig:sg_gan_vs_ours}), or that there is a motorcyclist on the bike (3rd row, \cref{fig:sg_gan_vs_ours}).

Finally, \methodName is easier to use for downstream tasks, as our method does not rely on  DensePose detections.
For example, keypoints are easier to edit for interactive editing applications.
Furthermore, detecting DensePose is challenging and unreliable for long-range detection, restricting its use in many scenarios (\eg anonymizing pedestrians on the street).
See \appSGGANAnonymizationComparison for examples of failure cases.

\begin{table}[t]
    \caption{
        Quantitative comparison of SG-GAN \cite{Hukkelas2022a} \vs ours.
        }
    \label{tab:sggan_vs_ours}
    \centering
    \begin{adjustbox}{max width=\linewidth}
    \begin{tabular}{lcccc}
        \toprule
        Method & FID $\downarrow$ & \fidC $\downarrow$ & PPL $\downarrow$ & OKS $\uparrow$ \\
        \midrule
        SG-GAN \cite{Hukkelas2022a} & 1.97 & 1.25 & 70.2 & 0.950\\
        \methodName (ours) & \textbf{1.68} & \textbf{0.43} & \textbf{47.8} & \textbf{0.972} \\
        \bottomrule
    \end{tabular}
    \end{adjustbox}
\end{table}

\section{Editability of \methodName}

StyleGAN \cite{Karras2018} is known for its disentangled latent space, and it is widely used for user-guided image editing, such as modifying images through text prompts \cite{Kocasari2022stylemc}.
However, most methods for editing images focus on unconditional GANs (or class-conditional GANs), and their application to image inpainting is less explored.
StyleMC \cite{Kocasari2022stylemc} is effective for editing faces with inpainting methods \cite{Hukkelas2022a}, but the same study finds editing human figures in-the-wild much harder \cite{Hukkelas2022a}.
We believe this limitation originates from the DensePose condition,  where descriptive conditions can be correlated with specific attributes.
This narrows the sampling probability,  which makes it harder to find meaningful directions for randomly sampled images.

\begin{figure}
    \includegraphics[width=.99\columnwidth]{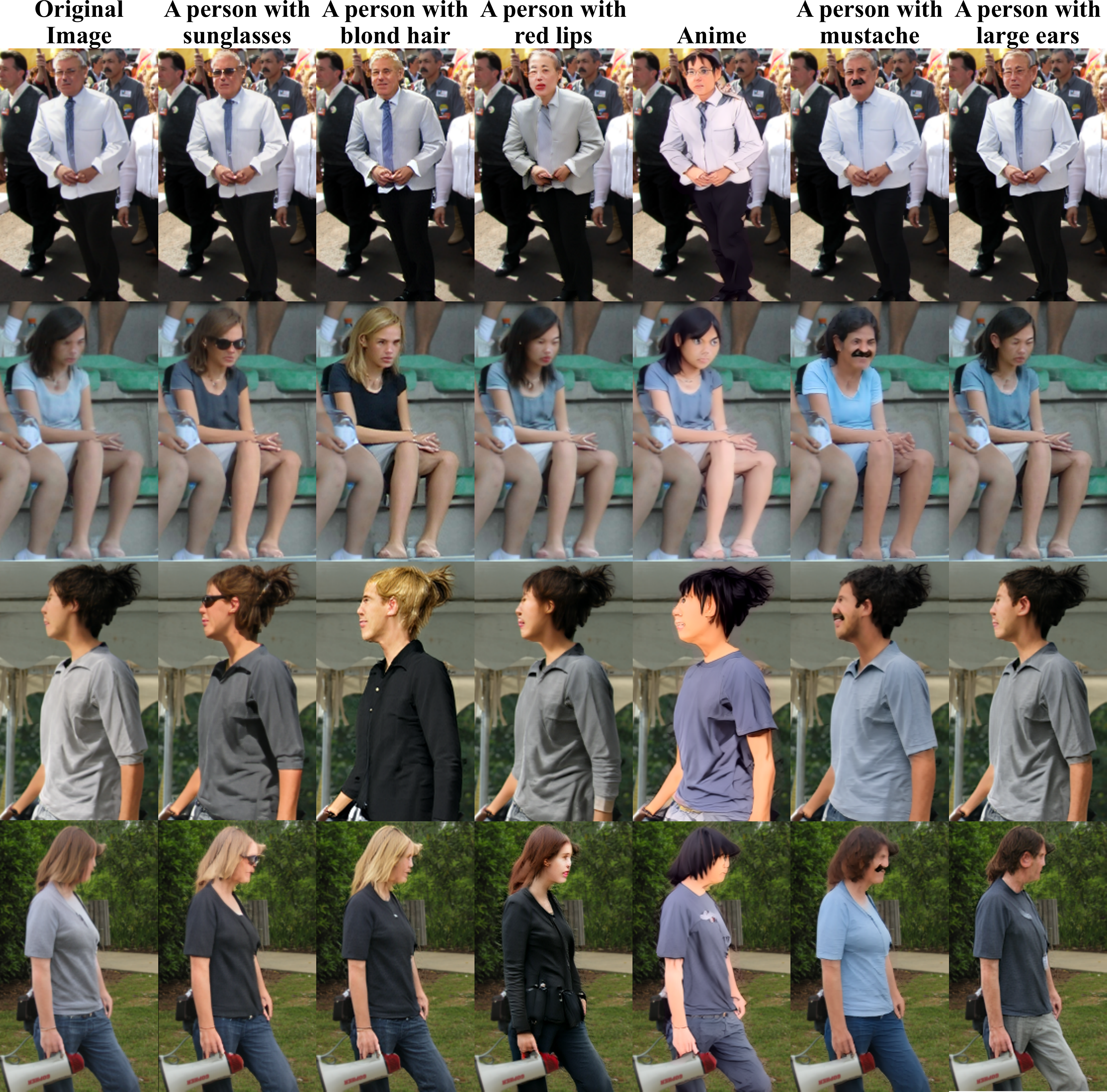}
    \caption{
        StyleMC \cite{Kocasari2022stylemc} edits with \methodName, where a global direction (from text prompt above each column) is added to the style code of the original (leftmost) image.
    }
    \label{fig:stylemc}
\end{figure}

\Cref{fig:stylemc} demonstrate that StyleMC \cite{Kocasari2022stylemc} is effective with \methodName to find semantically meaningful directions in the GAN latent space.
StyleMC finds global directions by manipulating random images towards a text prompt using a CLIP encoder \cite{Radford2021CLIP}, where the directions are found over 1280 images.
We find that StyleMC combined with \methodName can edit a wide range of attributes, even quite specific attributes such as the size of the ears.
However, we do note that editing some attributes results in changes to other correlated attributes.
For example, the edit "blond hair" induces slight changes to the skin color.
Furthermore, some attributes are more challenging to edit.
For instance, introducing "red lips" to a body inferred as a male can result in significant semantic changes (top row, \cref{fig:stylemc}).
It is unclear whether this limitation is a result from the editing technique or TriA-GAN itself.
We believe these correlations are inherent in the training datasets of CLIP or TriA-GAN.

\section{Conclusion}
\methodName has enabled the generation of human figures in any desirable pose and location given a sparse set of keypoints, resulting in a new state-of-the-art for person synthesis on the FDH dataset.
Key to our method is leveraging pre-trained feature networks for the discriminator.
We demonstrate that a carefully designed training strategy combined with feature networks suited to discriminate human figures substantially improves synthesis quality.
\methodName is the first to demonstrate reliable attribute editing of human figures via text prompts, which we believe will be highly practical for many applications.

\paragraph{Societal Impact}
Synthesizing human figures has a range of useful applications everywhere, from content creation to anonymization purposes.
However, similar to all learning-based generative models, the synthesized human figures adhere to the sampling probability of the dataset.
In our case, the dataset originates from Flickr, which means that our generator follows its biases and is less likely to synthesize people from underrepresented groups on the website.
Furthermore, our work focuses on generating lifelike humans, which carries the potential for abuse (\eg DeepFakes).
We note that the community has made a concerted effort to address this issue, through initiatives like the DeepFake Detection Challenge \cite{Dolhansky2020}, or embedding watermarks into images from generative models \cite{Yu2021}.

\begin{figure}
    \includegraphics[width=\columnwidth]{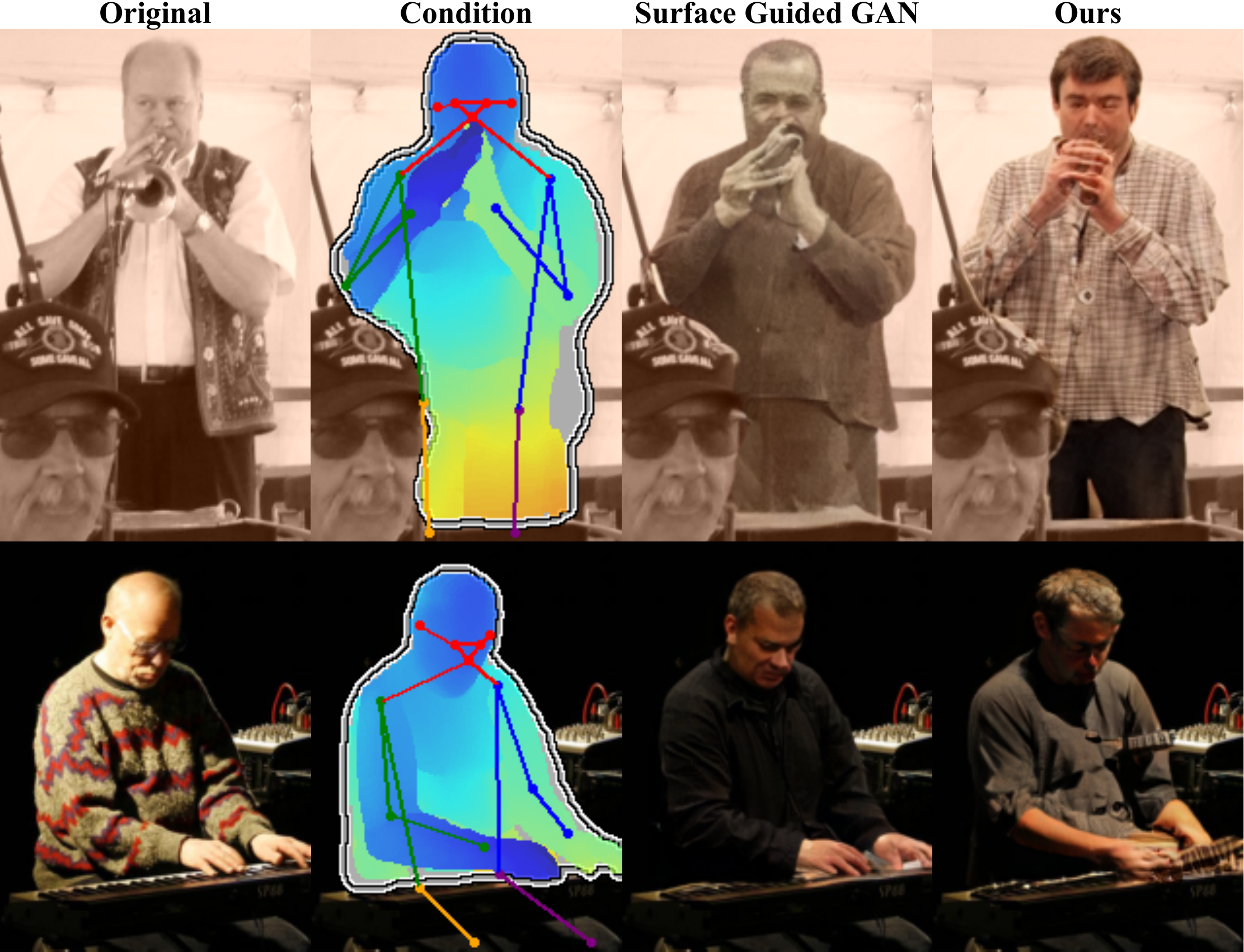}
    \caption{Failure cases of \methodName.}
    \label{fig:failure_cases}
\end{figure}

\subsection{Limitations}
\methodName sets a new state-of-the-art for human figure synthesis in-the-wild.
Exploring methods for disentangling the latent space from the pose, body shape, and environment are exciting future avenues.
Currently, the sampling space of \methodName is highly dependent on the conditional information, where it can collapse into a single synthesized identity given certain conditions.
Disentangled person image generation can mitigate this, by disentangle pose, appearance, and context.
However, current methods require datasets with paired images \cite{Sun2018person,Sarkar2021}, which are less diverse and small.

The key limitation of \methodName is handling more complex interactions with objects (\cref{fig:failure_cases}).
This is particularly true for generating realistic hands/fingers, \eg when playing the piano.
SG-GAN \cite{Hukkelas2022a} often improve on \methodName in such scenarios if the DensePose information explicitly describes the interaction.
But, it still struggles in cases where it is not clear (\eg playing the masked-out trumpet).

\methodName is hard to edit for attributes that are less frequent in the FDH dataset.
For example, many images do not contain the lower body and attempting to find editing directions for "a person wearing red pants" results in editing other attributes as well.
Whether this is a limitation to the editing method, or \methodName is an open question.

\paragraph{Acknowledgement}
The computations were performed on the NTNU IDUN/EPIC computing cluster \cite{sjalander2019epic} and the Tensor-GPU project led by Prof. Anne C. Elster.
Furthermore, we thank Rudolf Mester for his general support and thoughtful discussion.%

{\small
\bibliographystyle{ieee_fullname}
\bibliography{main}
}

\end{document}